\documentclass[letterpaper, 10 pt, conference]{ieeeconf}  %

\IEEEoverridecommandlockouts                              %

\usepackage{amsmath}
\usepackage{amssymb}
\usepackage{pdfpages}
\usepackage{xcolor}%
\usepackage{colortbl}
\usepackage{threeparttable}
\usepackage{multirow}
\usepackage{graphics} %
\usepackage{hyperref}
\usepackage{float}
\usepackage{todonotes}
\usepackage{cleveref}
\usepackage{pifont}
\usepackage[export]{adjustbox}
\usepackage{caption}
\usepackage{subcaption}

\usepackage{color,soul}

\title{\Large \bf
Cross-Modal Analysis of Human Detection for Robotics: An Industrial Case Study
}

\author{Timm Linder, Narunas Vaskevicius, Robert Schirmer and Kai O. Arras%
\thanks{This work has received funding from the European Union's Horizon
2020 research \& innovation programme under grant agreement 732737
(ILIAD) and 101017274 (DARKO).}%
\thanks{All authors are with Robert Bosch GmbH, Corporate Research, 71272 Renningen, Germany. 
        {\tt\small firstname.lastname@de.bosch.com}}%
}

\def\rect{{$\blacksquare \! \blacksquare$}} 
\def\yolo{{YOLO\,v3 (na\"ive depth) }} 

\definecolor{color:arras}{HTML}{ff6347}
\definecolor{color:drow-spaam}{HTML}{d2691e}
\definecolor{color:drspaam}{HTML}{ffab00}
\definecolor{color:leigh}{HTML}{d70012}
\definecolor{color:parta2-kitti}{HTML}{0a9c7f}
\definecolor{color:point-pillars}{HTML}{245525}
\definecolor{color:secfpn-dv-iliad}{HTML}{00ff00}
\definecolor{color:secfpn-dv-kitti}{HTML}{00bf00}
\definecolor{color:secfpn-kitti}{HTML}{008600}
\definecolor{color:uol-svm}{HTML}{a49d0a}
\definecolor{color:mobility-aids-2017-rgb}{HTML}{1b64b9}
\definecolor{color:naive-YOLOv3}{HTML}{6720f4}
\definecolor{color:pcl-roi}{HTML}{38b9f1}
\definecolor{color:pcl-roi-hog-svm}{HTML}{38b9f1}
\definecolor{color:rgbd-YOLOv3}{HTML}{f013e1}
\definecolor{color:rgbd-pose3d}{HTML}{0000FF}

\definecolor{color:fusion2}{HTML}{55ff55}
\definecolor{color:fusion3}{HTML}{ffcc00}
\definecolor{color:fusion9}{HTML}{ff80e5}

\definecolor{badcol}{HTML}{FFD4CC}
\definecolor{mehcol}{HTML}{FFFACF}  %
\definecolor{nicecol}{HTML}{C0F5B8} %
\def\nicecell{\cellcolor{nicecol}} 
\def\mehcell{\cellcolor{mehcol}}
\def\badcell{\cellcolor{badcol}}

\newcommand\copyrighttext{%
	\footnotesize \textcopyright 2021 IEEE. Personal use of this material is permitted.
	Permission from IEEE must be obtained for all other uses, in any current or future
	media, including reprinting/republishing this material for advertising or promotional
	purposes, creating new collective works, for resale or redistribution to servers or
	lists, or reuse of any copyrighted component of this work in other works.
}
\newcommand\copyrightnotice{%
	\begin{tikzpicture}[remember picture,overlay]
	\node[anchor=north,yshift=-10pt] at (current page.north) {\fbox{\parbox{\dimexpr\textwidth-\fboxsep-\fboxrule\relax}{\copyrighttext}}};
	\end{tikzpicture}%
}

\begin{document}

\maketitle
\copyrightnotice

\thispagestyle{empty}
\pagestyle{empty}

\begin{abstract}
Advances in sensing and learning algorithms have led to increasingly mature solutions for human detection by robots, particularly in selected use-cases such as pedestrian detection for self-driving cars or close-range person detection in consumer settings. Despite this progress, the simple question \emph{which sensor-algorithm combination is best suited for a person detection task at hand?} remains hard to answer.
In this paper, we tackle this issue by conducting a systematic cross-modal analysis of sensor-algorithm combinations typically used in robotics. We compare the performance of state-of-the-art person detectors for 2D range data, 3D lidar, and RGB-D data as well as selected combinations thereof in a challenging industrial use-case. %

We further address the related problems of data scarcity in the industrial target domain, and that recent research on human detection in 3D point clouds has mostly focused on autonomous driving scenarios. To leverage these methodological advances for robotics applications, we utilize a simple, yet effective multi-sensor transfer learning strategy by extending a strong image-based \mbox{RGB-D} detector to provide cross-modal supervision for lidar detectors in the form of weak 3D bounding box labels.

Our results show a large variance among the different %
approaches 
in terms of detection performance, generalization, frame rates and computational requirements.
As our use-case contains difficulties representative for a wide range of service robot applications,
we believe that these results point to relevant open challenges for further research and provide valuable support to practitioners for the design of their robot system.

\end{abstract}

\section{Introduction}
\label{sec:introduction}
The ability of robots to detect people in their vicinity is of great importance in consumer, industrial and automotive application domains. Use-cases range from pedestrian prediction for driverless cars, user recognition for consumer robots, to safety-critical operator detection for collaborative manipulators. They vary in  requirements and constraints -- examples include minimal accuracy, cost of false positives, computational budget, privacy constraints or maximal system cost --, and it comes as no surprise that solutions differ strongly across and within application domains and communities. This variety means that

\begin{itemize}
\item[1.] Progress in one domain cannot be readily transferred to another due to domain gap issues (e.g. indoor/outdoor, different sensor resolutions or mounting positions).

\item[2.] The availability of annotated data is very domain- and sensor-specific and far from uniform (e.g. plenty of image data sets for autonomous driving vs. few 3D data sets in typical service robot scenarios). This makes detector retraining and comparisons of approaches across sensory modalities and domains a difficult task.

\item[3.] With a person detection task at hand, characterized by such requirements and constraints, it is hard to navigate the large space of sensor-algorithm combinations and identify the most promising options. 
\end{itemize}

\begin{figure}[t!]
    \centering
    \includegraphics[width=\columnwidth]{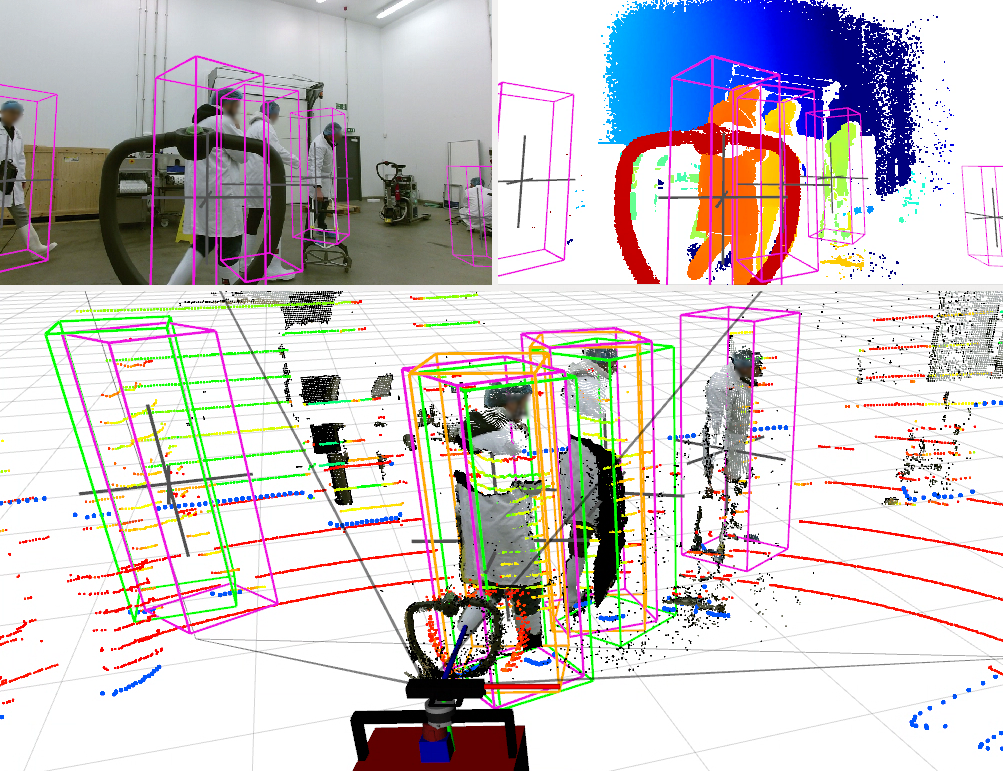}
    \caption{Which sensor-algorithm combination is best suited for a person detection task at hand? Example scene from our data set seen by an RGB-D sensor (RGB and color-encoded depth image in first row + colored point cloud), a 2D safety laser (blue points) and a 3D lidar (points in rainbow colors). Boxes are detections in different modalities (see \autoref{fig:wp3-det-comparison-quantitative}), grey crosses visualize groundtruth centroids.}
    \label{fig:datasets:iliad-dataset-challenges}
\end{figure}

The latter point is of large practical importance. Given, for example, a cost-sensitive consumer robot without GPU, how do state-of-the-art detectors for low-cost range cameras and 2D lidars compare relative to the required accuracy, or do classical CPU-based machine learning methods suffice? Or given a hospital delivery robot, does the gain in privacy from using a 3D lidar over an RGB-D camera justify the higher sensor cost and drop in detection performance?

To address those issues within a robotics scope, this paper makes the following contributions:

We conduct a cross-modal analysis of sensor-algorithm combinations typically used in robotics. We compare the performance of state-of-the-art person detectors for the  commonly used 2D lidar, 3D lidar and RGB-D sensors as well as selected combinations thereof. For each modality, we also consider a classical GPU-free learning baseline. Although a large body of literature on multi-model sensor fusion for human detection exists, to the best of our knowledge, a comparison of this breadth has not been done before.

 Our second contribution addresses points 1 and 2:
 We extend our RGB-D YOLO~\cite{Linder2020ICRA} method capable of real-time detection of human 3D centroids in RGB-D data to regress 3D oriented bounding boxes. We learn 3D bounding box estimation in RGB-D YOLO solely from synthetic data to obtain an accurate 3D human detector. We then demonstrate that such a detector trained without manual 3D annotations can be utilized to transfer knowledge across sensors and improve  detectors on the target domain via weak supervision.

For evaluation, we collected data in a warehouse using a mobile platform equipped with the mentioned sensors. The environment contains difficulties typical for service robot use-cases: people in varying densities and body poses, in narrow and open spaces, in proximity and interaction with objects and walls, uniform clothing, and in occlusion from foreground objects and other people, see Figs. \ref{fig:sensor-setup} and \ref{fig:wp3-det-comparison-quantitative}.

\section{Related work}

There is a large body of literature on human detection in robotics and related fields whose discussion is beyond the scope of this paper, see e.g. the surveys
\cite{Bellotto2018,Ciaparrone2019}. %
We discuss methods that are part of our evaluation in \autoref{section:evaluated-approaches}.
Multi-modal approaches such as  \cite{Bellotto2009,Linder2016,Volkhardt2013,Qi2018,ku2018}  typically combine two or more sensors with a focus on sensor/detector fusion, performance maximization or domain adaptation. Here, on the other hand, we strive for an unbiased cross-modal comparison systematically covering relevant modalities and baselines for robotics. %

Existing benchmarks and datasets only partly support this goal. Either because they are single-sensor datasets \cite{luber2011,Weinrich2014,Beyer2018} or due to domain gap issues mentioned in \autoref{sec:introduction}. Large-scale automotive benchmarks such as KITTI \cite{Geiger2012}, Waymo Open \cite{Sun2020CVPR} or NuScenes \cite{Caesar2020CVPR} consider sensor setups and operating conditions different from typical service robot use-cases. %
An exception is JackRabbot \cite{MartinMartin2019}, a multi-modal robotic benchmark and dataset collected on a university campus indoors and outdoors. As here we focus on industrial use-cases we do not consider this interesting dataset and leave its inclusion to concurrent \cite{jia2021domaingaps} and future work. 

Creating domain specific datasets with manual 3D annotations for supervised learning requires extensive resources. Transfer learning is one of the tools to reduce this effort and address the \emph{domain adaptation} challenge. While it can take a variety of forms (e.g. sim-to-real), in this work we focus on transferring real-world knowledge across sensors. The domain gap exists even across different sensor models of the same modality and has been addressed in prior works \cite{rist2019iv}, \cite{piewak2019itsc} where 3D lidar-to-lidar supervision is used for object detection and semantic segmentation tasks. In contrast to these methods, we investigate transfer learning across different modalities. \cite{piewak2018eccv} and \cite{wang2019ral} apply cross-modal transfer from RGB to 3D lidar, but for semantic and instance segmentation, as opposed to 3D human detection considered in this paper. Our strategy is inspired  by  previous work on multi-sensor transfer learning \cite{Yan2018}. However, in our case, we use a very strong, deep learning-based RGB-D detector as teacher network, making the entire pipeline much simpler as no complex tracking and outlier rejection is required.

\section{Evaluated modalities and approaches}
\label{section:evaluated-approaches}

In the following, we provide an overview of the human detection approaches that are part of our comparison, grouped by sensor modality.

\subsection{2D laser detectors}

Many service robots are equipped with 2D safety laser scanners at floor level for localization, obstacle avoidance and safety. A recent approach to person detection in 2D range data has been presented by Beyer \emph{et al.} \cite{Beyer2018}, which %
 introduced a novel CNN-based method called \emph{DROW3x} that can fuse information from temporally consecutive frames to recognize typical human leg motion. On a dataset from an elderly care facility, the method clearly outperformed several existing approaches that rely on handcrafted geometric features and classical machine learning techniques (random forests or AdaBoost). However, the latter methods have an advantage in terms of hardware requirements, since they are able to run efficiently on a CPU. Therefore, besides \emph{DROW3x}, we also include the classical methods by \emph{Arras et al.}~\cite{Arras2007} and \emph{Leigh et al.}~\cite{Leigh2015} in our comparison. The latter follows a similar approach as \cite{Arras2007} but detects individual legs and associates them over time using a simple Kalman filter-based tracker. While not strictly a detection-only approach, we treat it here as such and rely on its data association stage to obtain human centroids for evaluation (see sec.~\ref{sec:experimental-setup}).

Very recently, a more efficient version of DROW3x, dubbed \emph{DR-SPAAM} \cite{Jia2020DRSPAAM}, has been proposed, which we also consider in our evaluation. By replacing the expensive voting grid with a non-maximum suppression step and by using a recurrent update scheme instead of explicit odometry-based multi-scan alignment for temporal fusion, it achieves significantly higher frame rates and is suitable also for deployment on embedded platforms (e.g. Nvidia Jetson). %

\subsection{3D lidar-based approaches}

\label{section:human-detection-lidar}

For 3D lidar-based human detection we use the approach by Yan et al.~\cite{Yan2018} as a classical machine learning baseline which does not require accelerator hardware for real-time performance. It uses an SVM classifier with handcrafted geometric features after proposal generation through Euclidean clustering.

The deep learning-based 3D approaches in our comparison have so far been evaluated mostly in autonomous driving use-cases. We selected methods based on their suitability for our use-case by taking into account their pedestrian detection performance in the KITTI benchmark \cite{Geiger2012} and their computational requirements. To facilitate integration and avoid strong implementation bias, we used the best-performing methods available at the time of our experiments within the MMDetection3D framework \cite{MMDetection3D}: 
\emph{SECOND} \cite{yan2018sensors}, \emph{SECOND with Dynamic Voxelization} \cite{Zhou2019endtoend},
\emph{PointPillars} \cite{Lang2019}, and \emph{PartA2-Net} \cite{shi2020part}.

 \emph{SECOND} %
 performs voxelwise feature extraction as in VoxelNet \cite{Zhou2018}, before feeding the intermediate representation into a sparse convolutional middle extractor that converts sparse 3D data into a 2D birds-eye view image. Finally, an SSD-like head \cite{Liu2016} outputs oriented 3D bounding boxes. \emph{SECOND with Dynamic Voxelization} overcomes the possible information loss due to stochastic point dropout in fixed voxelizations and yields deterministic voxel embeddings and more stable detection outcomes. In AD scenarios, the method achieves higher accuracy at approximately the same runtime cost as \cite{yan2018sensors}. \emph{PointPillars} is a faster variant of SECOND with a relatively small GPU memory footprint. It replaces 3D convolutional layers with dense 2D convolutions by learning a representation of point columns (pillars). \emph{PartA2-Net} consists of two stages. The first, part-aware stage generates 3D proposals with corresponding point-wise features and predicted object point locations relative to the oriented 3D bounding box. This information is aggregated in the second part-aggregation stage to retrieve final object detections and refined 3D bounding boxes.

\subsection{RGB-D methods}

Our recent work \cite{Linder2020ICRA} investigated the task of 3D human detection using data from a time-of-flight RGB-D camera for robotics applications. It provides a comprehensive overview of the state-of-the-art approaches and compares them on an intralogistics dataset, which is a subset of the data used in this evaluation.
In the same work we proposed an image-based detection approach, \emph{RGB-D YOLO}, which extends the YOLO\,v3 architecture with a 3D centroid loss and mid-level feature fusion to exploit complementary information from both RGB and depth modalities. We also introduced an efficient transfer learning strategy to benefit from both pre-training on large-scale 2D image datasets such as MS COCO~\cite{COCO14}, and highly randomized synthetic RGB-D data with accurate 3D groundtruth \cite{Linder2019}. The experiments showed that the proposed method achieves higher detection accuracy in 3D space than state-of-the-art baselines.

In this study, besides \emph{RGB-D YOLO}, we include further top performing baselines from \cite{Linder2020ICRA}. A na\"ive image-based method \cite{Linder2018} lifts detected 2D bounding boxes into 3D space by sampling depth values within the bounding box from the registered depth image. While such an approach can yield good results in simpler cases, there exist cases where e.g. the majority of all valid depth pixels inside the 2D bounding box belong to entirely different foreground or background objects, leading to wrong depth estimates. As a CPU-only method we chose the work by Munaro et al.~\cite{Munaro2014}, which combines HOG features with an SVM classifier after performing a head-based subclustering in the point cloud. It performed reasonably well in comparison to deep learning methods on the benchmark in~\cite{Linder2020ICRA}. \emph{Mobility Aids}~\cite{Vasquez2017} applies Euclidean clustering on the 3D point cloud for proposal generation, followed by a CNN-based classifier on the RGB or depth images.
\emph{RGB-D Pose 3D} \cite{Zimmermann2018} predicts 2D human keypoints in a bottom-up fashion using OpenPose \cite{Cao2017}, then lifts the body into 3D space by retrieving depth values at pre-defined body joint locations (e.g. neck), before centering a 3D voxel occupancy grid on the body to predict 3D joint locations. Both deep learning-based methods \cite{Vasquez2017} and \cite{Zimmermann2018} utilize a geometric 3D point cloud representation, which comes with certain drawbacks, e.g., the 3D stage could fail to detect objects in locally sparse point clouds.

\section{Multi-sensor Transfer Learning}

\begin{figure}
    \vspace*{1ex}
	\centering
	\includegraphics[width=1.0\columnwidth]{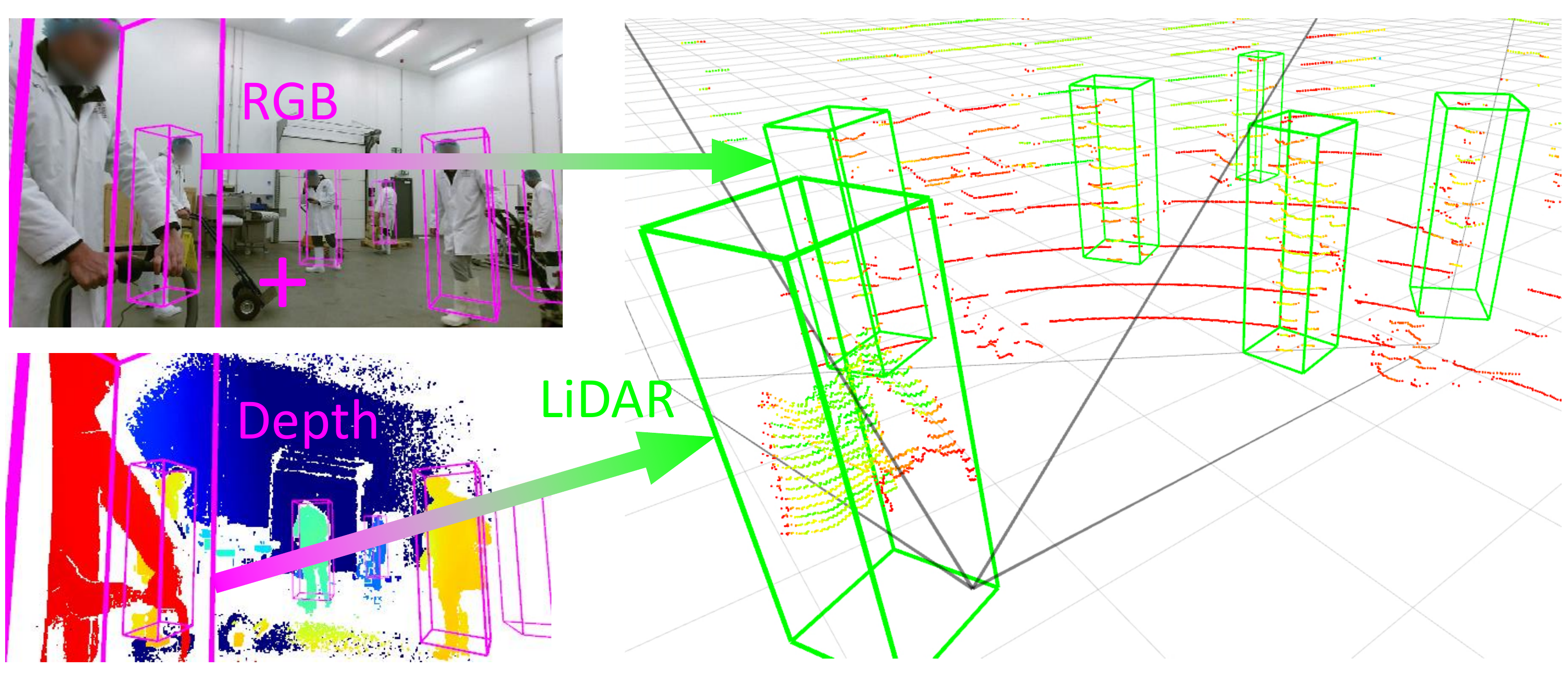}	
	\caption{Multi-sensor transfer learning by using the RGB-D YOLO detector, extended to provide oriented 3D bounding boxes, for learning lidar-based human detection in the target domain without extensive need for manual 3D labeling.%
	} %
	\label{fig:transfer-learning-lidar-scenes}
\end{figure}

Our initial experiments indicated that the 3D LiDAR-based detectors, when trained on automotive datasets \cite{Geiger2012}, would significantly underperform in certain scenarios compared to 2D laser-based methods, which is unexpected due to their significantly higher resolution in the vertical dimension. The most likely reason for this is the domain gap that results from different sensor resolutions, mounting heights, and our focus on indoor environments, which we aim to address by retraining them on the target domain. However, manually labeling 3D point clouds with 3D oriented bounding boxes (3D OBB) is time-consuming. Given the availability of approximately time-synchronized \mbox{RGB-D} frames from a time-of-flight camera like the Kinect v2, we instead propose to exploit a strong \mbox{RGB-D} detector to provide cross-modal supervision in the form of weak groundtruth box labels, as indicated in \autoref{fig:transfer-learning-lidar-scenes}. Unlike other approaches \cite{Yan2018,jia2020selfsupervised}, we do not rely on any additional tracking or explicit outlier rejection stage and do not use 2D bounding boxes as an intermediate representation with resulting information loss.

\subsection{Extension of RGB-D YOLO for 3D OBB supervision}

\begin{figure}[t]
	\centering
	\includegraphics[width=1.0\linewidth]{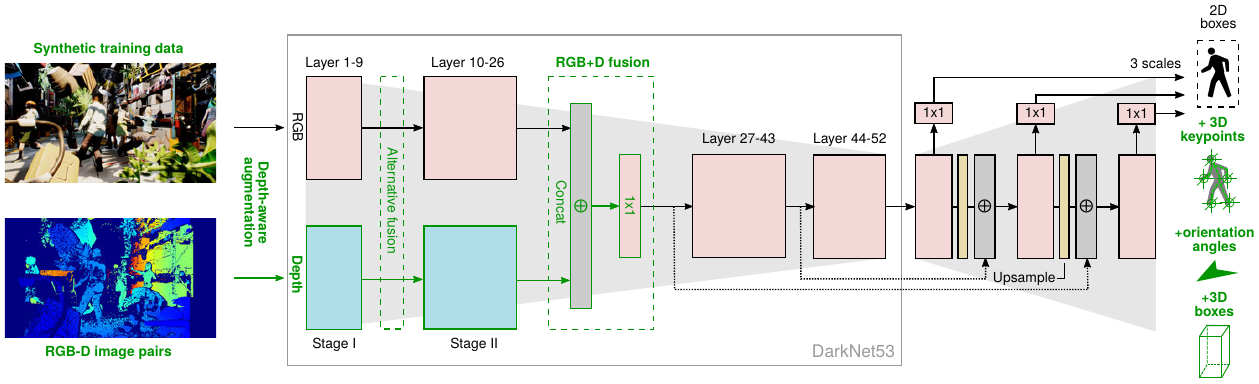}
	\vspace*{-3ex}
	\caption{Extended RGB-D YOLO architecture with regression of 3D body keypoints including pelvis joints, orientation angles and 3D bounding boxes.}
	\label{fig:rgbd-yolo-arch-extended}
\end{figure}

To this end, as shown in Figure~\ref{fig:rgbd-yolo-arch-extended}, we extend our \mbox{RGB-D} YOLO\,v3 approach to regress not only human centroids (pelvis joints), but also oriented 3D bounding boxes $(x,y,z,d,w,h,\theta)$ with centroids $(x,y,z)$, extents $(d,w,h)$ and yaw angles~$\theta$. For 3D box centroids, we use the same 2.5D keypoint loss with normalized pixel coordinates $u, v$ and metric depth $z$ as in \cite{Linder2020ICRA}, in addition to the regular 2D box center, scale and objectness losses, while directly regressing metric 3D extents using an $l_1$ loss. For the yaw angle, we use a biternion representation $(\mathrm{cos}~\theta,~ \mathrm{sin}~\theta)$ as in \cite{Beyer2015,Lewandowski2019a} with a cosine similarity loss. The resulting combined loss term for training is then
\begin{equation}
    \mathcal{L} = \mathcal{L}_\mathrm{centroids} + \mathcal{L}_\mathrm{angle} + \mathcal{L}_\mathrm{extents} +
    \mathcal{L}_\mathrm{2D}.
\end{equation}

To obtain groundtruth 3D box labels in RGB-D, we rely on our synthetic training data generation pipeline \cite{Linder2019} and learn this aspect solely from synthetic data.

\subsection{Transfer learning approach}

\label{section:transfer-learning-from-rgbd}

\begin{figure}
	\centering
	\includegraphics[width=1.0\linewidth]{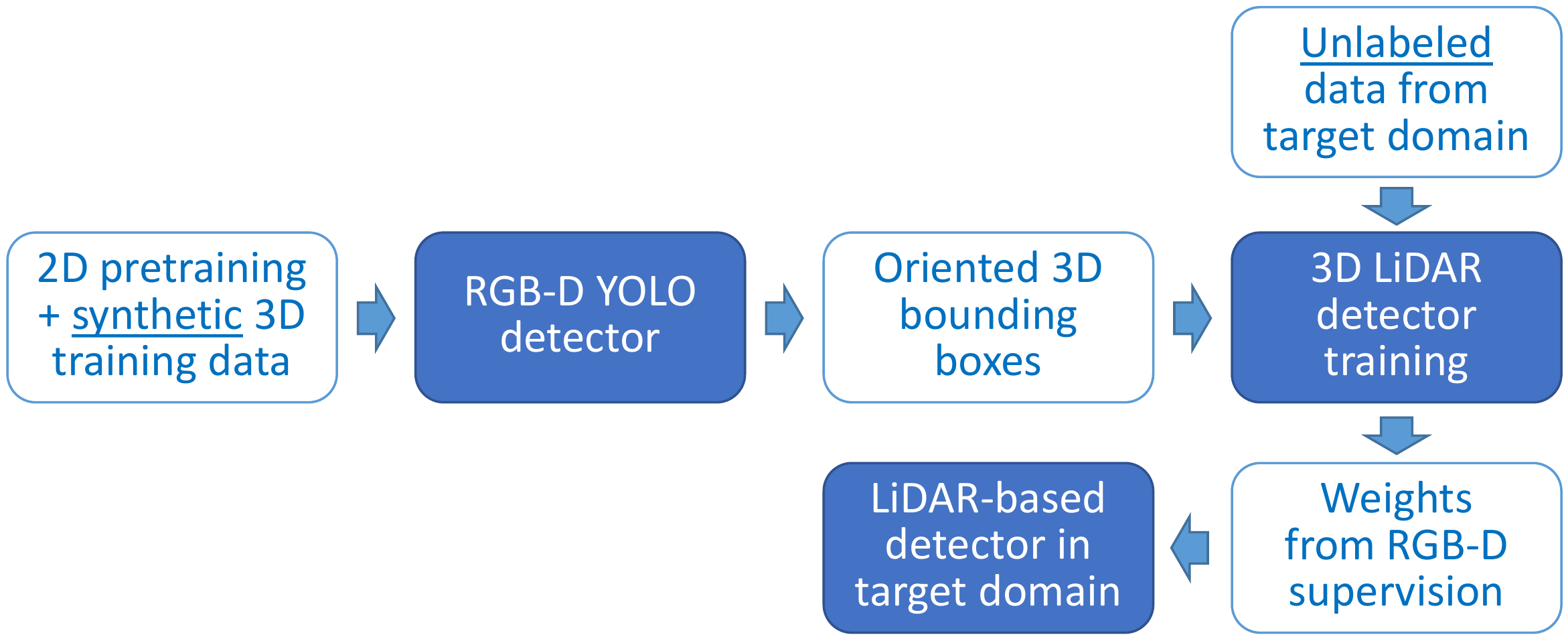}
	\vspace*{-3ex}
	\caption{Pipeline for weakly supervised 3D lidar detector training without need for manual annotations by using oriented 3D bounding boxes from the extended \mbox{RGB-D} YOLO detector (trained with synthetic data) as supervision signal.}
	\label{fig:transfer-learning-pipeline}
\end{figure}

We then directly transfer the 3D box labels from the extended RGB-D YOLO detector to lidar point clouds via the existing extrinsic sensor calibration. We do not include any additional tracking stage, which could induce false positives due to delayed track deletion. Since during lidar detector training, that we restrict to the Kinect v2 field of view and then extend it back to full 360 degrees using rotation augmentation, groundtruth boxes without any lidar points are discarded as positive samples, the few wrongly localized 3D boxes that our teacher network still produces often have no negative impact on the training process. \autoref{fig:transfer-learning-pipeline} provides an overview of the proposed training pipeline.

\section{Experiments}

\subsection{Industrial dataset from an intralogistics warehouse}

\begin{figure}
    \centering
    \includegraphics[width=0.85\columnwidth]{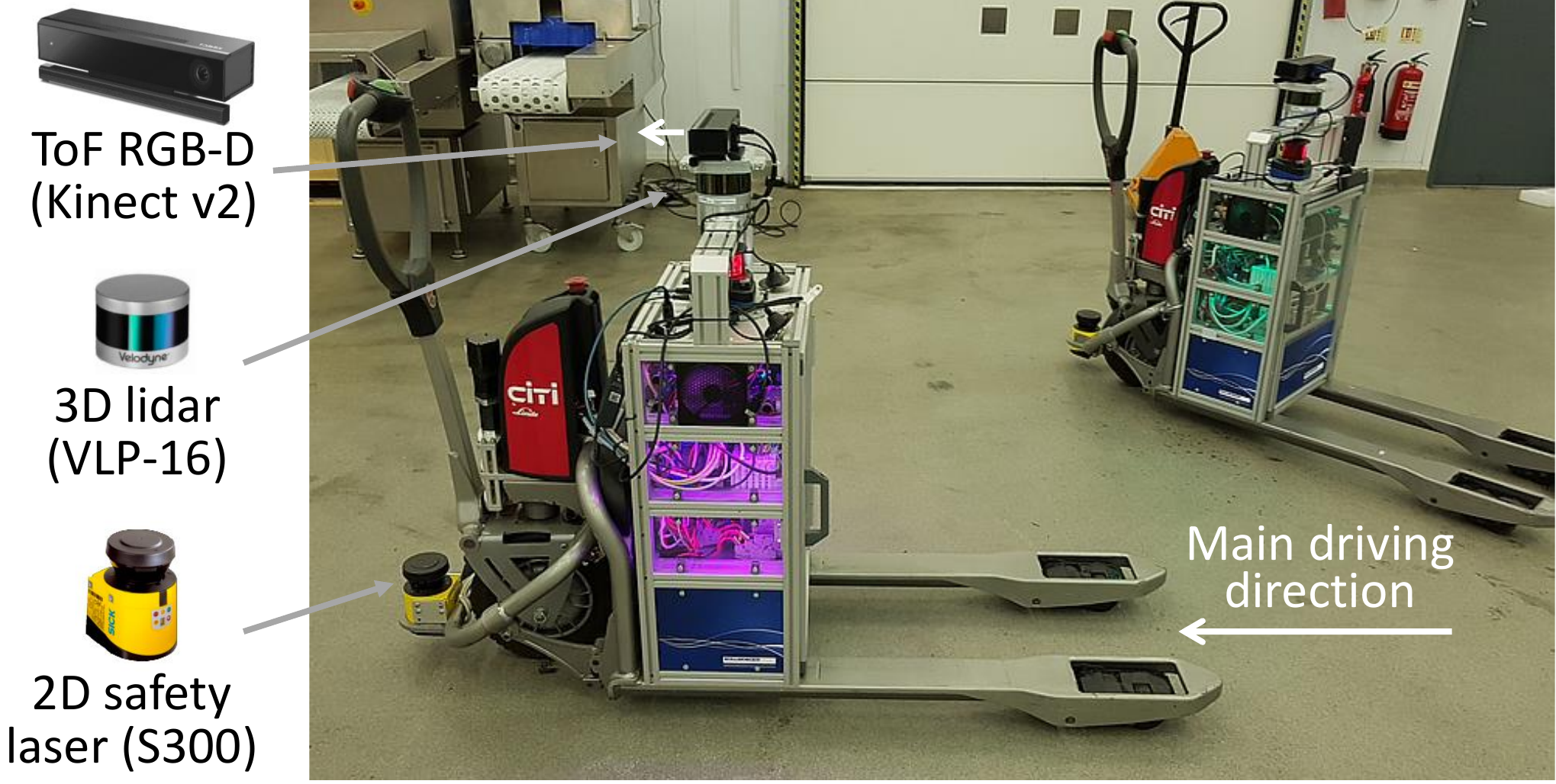}
    \caption{Multimodal sensor setup used in our evaluation}
    \label{fig:sensor-setup}
\end{figure}

To evaluate human detectors in industrial environments, we recorded over 55 hours of data with our mobile robots in different intralogistics scenarios over a timespan of several weeks. For the purpose of this first comparison, we hand-selected and manually labeled two challenging multimodal sequences, that are representative of commonly observed scenarios, with 3D person centroid trajectories. They span three minutes in total and originate from a robot within a food factory warehouse environment. Both scenes have been recorded using the sensor setup in Figure~\ref{fig:sensor-setup}. Sequence~(a), \emph{NCFM Dynamic}, has a higher person density with a larger number of temporary occlusions, e.g. due to persons pushing carts and pallet trucks, and has been recorded with a static sensor platform in an open space with a lot of highly dynamic motion and including also non-standing poses. Instead, Sequence (b), \emph{NCFM Storage Room}, was acquired with a moving robot in a narrow storage room with only 1--2 persons that are always in upright poses and often in close proximity to shelves, walls or the robot.

\subsection{Experimental setup}
\label{sec:experimental-setup}

We evaluate all previously described methods, that are representative of different types of approaches on the respective sensor modality, with regard to their detection performance in 3D space. We report \emph{average precision (AP)}, i.\,e., the area under the precision-recall curve over varying detection thresholds, and the \emph{peak-F1 score} at a single confidence threshold, which represents a reasonable compromise between detector precision and recall. To compute these metrics, we perform ground truth association using the estimated centroids over the ground plane, which is the most common representation to all approaches and modalities. For a fair comparison, we ignore centroid height above ground\footnote{Note that the 2D laser-based approaches do not output 3D bounding boxes and do not provide estimates of the centroid height over ground.}. We consider a person as correctly detected if a detection falls within 0.5~m of the closest ground truth centroid, discount ground truth annotations that are heavily occluded in the 3D lidar point cloud (containing less than 7 lidar returns), and only evaluate detections within the (most limited) field of view of the Kinect v2 \mbox{RGB-D} sensor, which is necessary for a fair comparison because the sensor modalities differ fundamentally in their horizontal field of view.

For each method, we use models trained on the dataset reported in parentheses behind the method name in \autoref{fig:wp3-det-comparison-quantitative}. For 3D lidar detector models trained on KITTI, we transform our input point clouds to compensate for different sensor mounting heights and normalization of lidar intensities in order to reduce the domain gap. For multi-sensor transfer learning on our ILIAD intralogistics dataset, we add unlabeled, temporally disjoint data from the same environment as the two evaluation sequences. For this, we sample around 27k additional frames in a KITTI-like format. For training, we adapt the maximum range to 15m as our point clouds contain no useful data beyond this distance.

To examine if detection results improve by (na\"ively) fusing detections of the best-performing detectors across different sensor modalities, we perform nearest-neighbor association of nearby detections from different detectors in Euclidean space with a gating distance of 0.5~m.

\section{Results and discussion}

\begin{figure*}
	\vskip-6mm
		\centering
		\begin{subfigure}[b]{0.45\linewidth}
			\centering
			\hspace*{-25mm}
			\begin{minipage}[b]{0.49\linewidth}
			\centering
			\includegraphics[height=3.7cm]{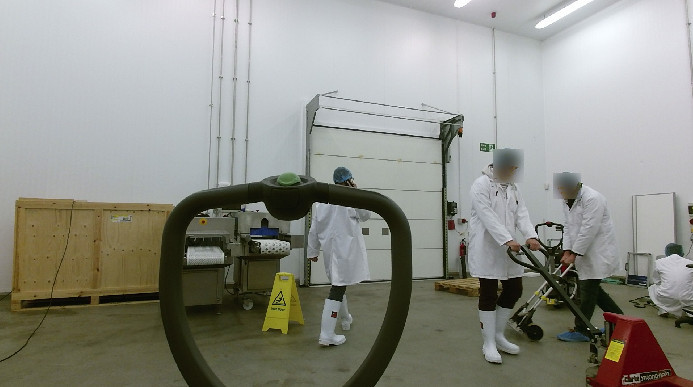}
			\includegraphics[trim=0 0mm 0 0mm, clip,width=60mm]{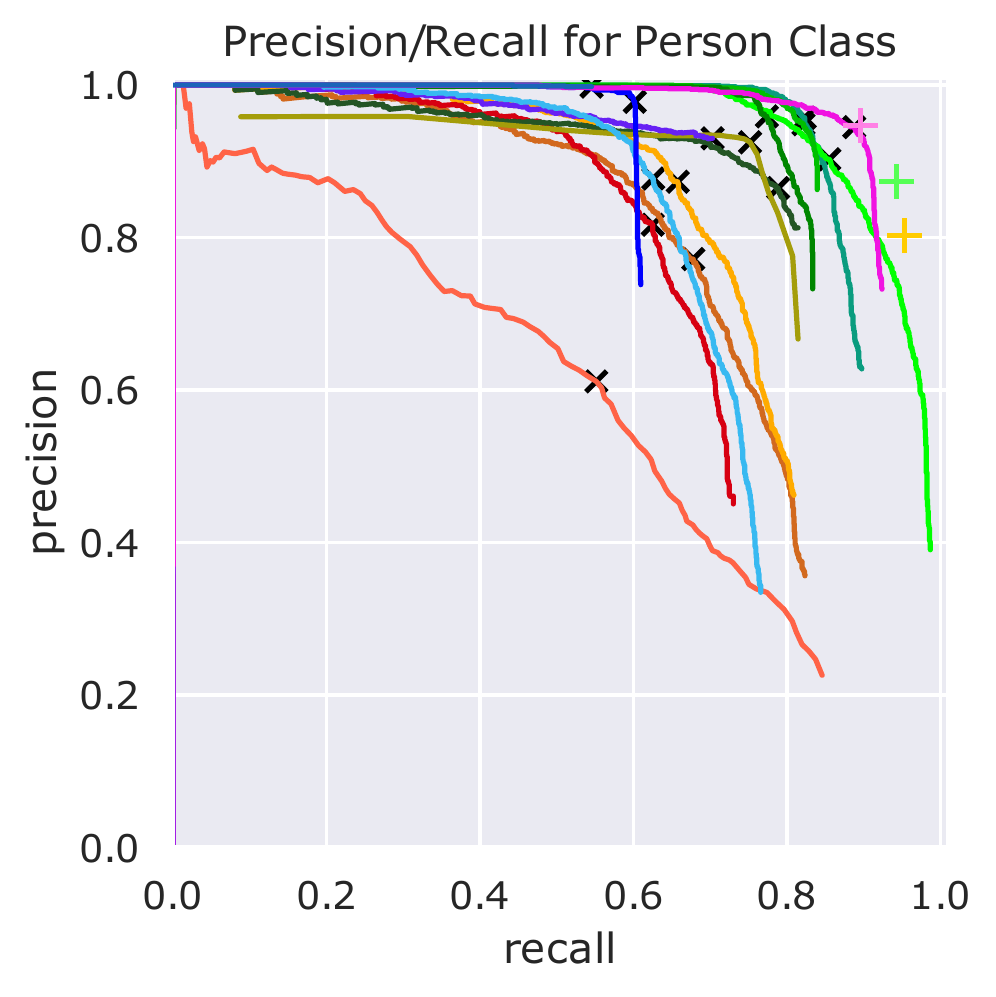}
			\end{minipage}
		    \caption{NCFM Dynamic scene}
		\end{subfigure}
        \hspace*{4mm}
		\begin{subfigure}[b]{0.45\linewidth}
			\centering
			\hspace*{-25mm}
			\begin{minipage}[b]{0.49\linewidth}
			\centering
			\includegraphics[height=3.7cm]{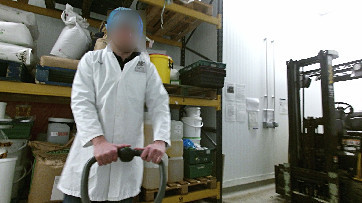}
			\includegraphics[trim=0 0mm 0 0mm, clip,width=60mm]{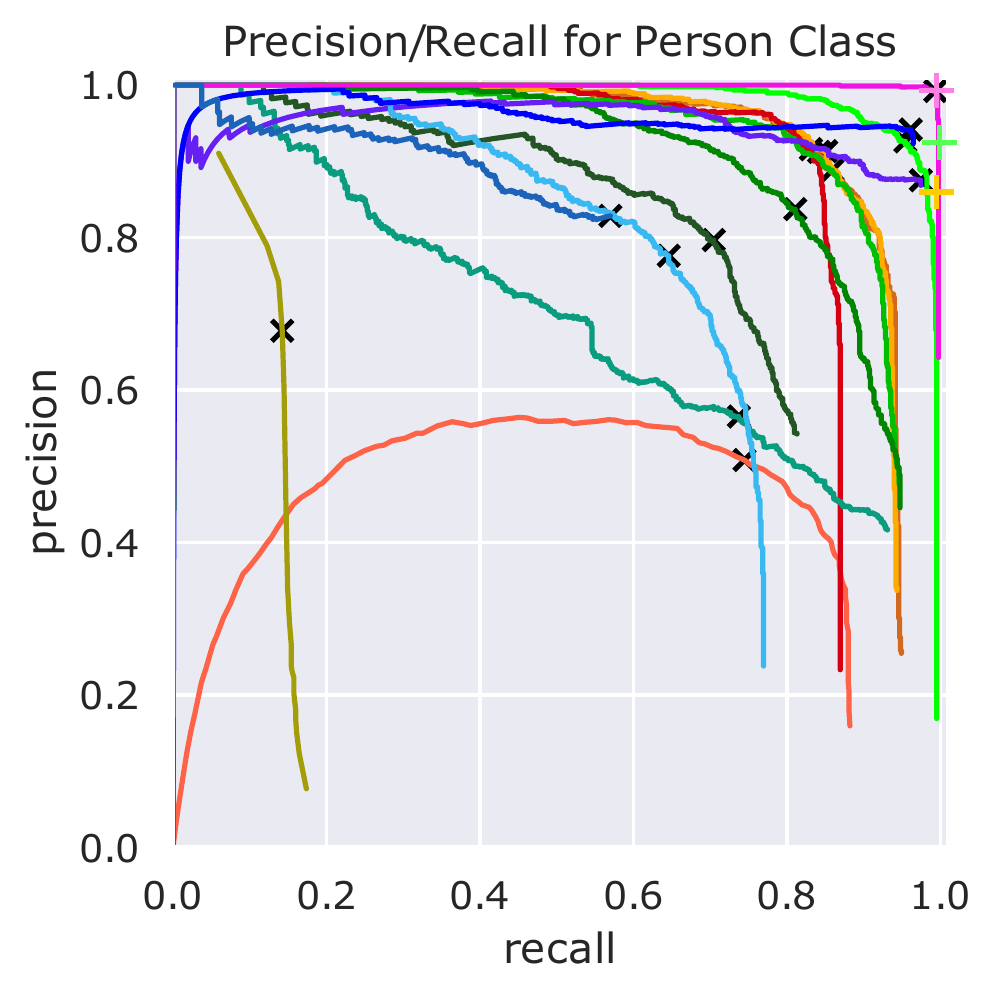}
			\end{minipage}
			\caption{NCFM Storage scene}
		\end{subfigure}
		
		\vspace*{4mm}
		\hspace*{-25mm}
		\resizebox{\linewidth}{!}{
			\input{tex/detector_comparison_defs}

\setlength\extrarowheight{4pt}
\begin{tabular}{|l|l|c|c|c|c|c|c|c|c|c|}
	\hline
	Modality & \begin{tabular}[c]{@{}l@{}}Method and training data \\ (if not specified: proprietary training set)\end{tabular} & \begin{tabular}[c]{@{}l@{}}Horiz. \\ FOV $^{\circ}$\end{tabular} & \begin{tabular}[c]{@{}l@{}}FPS\\ (Hz)\end{tabular} & \begin{tabular}[c]{@{}l@{}}VRAM\\ (MiB)\end{tabular} & \multicolumn{3}{c|}{ \begin{tabular}[c]{@{}l@{}}AP[0.5m] $\uparrow$\\ \hspace{2.5ex} in \%\end{tabular} } & \multicolumn{3}{c|}{ Peak-F1 $\uparrow$ } \\ %
	&&&&&
	(a) & (b) & (a)+(b) & (a) & (b) & (a)+(b) \\ \hline
	& \textcolor{color:arras}{\rect} Arras \cite{Arras2007} as in \cite{Beyer2018} (DROW) &  & 25 & - & 56.0 & 47.4 & 50.7 & 0.58 & 0.60 & 0.59 \\ \cline{2-2} \cline{4-11} 
	2D laser & \textcolor{color:leigh}{\rect} Leigh \cite{Leigh2015} as in \cite{Beyer2018} (DROW) & 180 & 22 & - & 67.4 & 84.6 & 72.2 & 0.71 & 0.87 & 0.75 \\ \cline{2-2} \cline{4-11} 
	& \textcolor{color:drow-spaam}{\rect} DROW3x, T=1 (DROW) \cite{Beyer2018} + NMS \cite{Jia2020DRSPAAM} &  & 49 & 827 & 72.7 & \underline{91.3} & 78.4 & 0.72 & 0.88 & 0.77 \\ \cline{2-2} \cline{4-11} %
	& \textcolor{color:drspaam}{\rect} DR-SPAAM (DROW) \cite{Jia2020DRSPAAM} &  & 48 & 829 & \underline{74.4} & 91.2 & \underline{79.8} & \underline{0.75} & \underline{0.88} & \underline{0.79} \\ \hline
	& \textcolor{color:uol-svm}{\rect} Object3D SVM Detector \cite{Yan2018} &  & 8 & - & 77.1 & 12.6 & 52.6 & 0.83 & 0.24 & 0.63 \\ \cline{2-2} \cline{4-11}
	& \textcolor{color:parta2-kitti}{\rect} PartA2-Net (KITTI) \cite{shi2020part} & & 7 & 5183 & 87.8 & 67.0 & 83.5 & \underline{0.89} & 0.64 & 0.79 \\ \cline{2-2} \cline{4-11}
	3D lidar & \textcolor{color:point-pillars}{\rect} PointPillars (KITTI) \cite{Lang2019} & 360 & 11 & 1217 & 77.5 & 72.8 & 76.3 & 0.83 & 0.75 & 0.80 \\ \cline{2-2} \cline{4-11}
	& \textcolor{color:secfpn-kitti}{\rect} SECOND (KITTI) \cite{yan2018sensors} & & 9 & 4999 & 82.3 & 87.5 & 84.1 & 0.86 & 0.82 & 0.85 \\ \cline{2-2} \cline{4-11} %
	& \textcolor{color:secfpn-dv-kitti}{\rect} SECOND-DV (KITTI) \cite{Zhou2019endtoend} & & 9 & 4979 & 82.9 & 90.4 & 86.0 &  0.88 & 0.87 & \underline{0.88} \\ \cline{2-2} \cline{4-11} %
	& \textcolor{color:secfpn-dv-iliad}{\rect} SECOND-DV (ILIAD), transfer learning & & 9 & 5025 & \underline{94.4} & \underline{97.9} & \underline{93.9} & 0.88 & \underline{0.94} & 0.87 \\ \hline %
	& \textcolor{color:pcl-roi-hog-svm}{\rect} PCL + HOG-SVM \cite{Munaro2014} & & 21 & - & 70.7 & 67.9 & 69.6 & 0.73 & 0.71 & 0.72 \\ \cline{2-2} \cline{4-11} %
	RGB-D & \textcolor{color:mobility-aids-2017-rgb}{\rect} Mobility Aids \cite{Vasquez2017} & & 23 & 1576 & 54.4 & 52.6 & 54.4 & 0.71 & 0.68 & 0.70 \\ \cline{2-2} \cline{4-11}
	(Kinect v2) & \textcolor{color:rgbd-pose3d}{\rect} RGB-D Pose 3D \cite{Zimmermann2018} & 86 & 1 & 4055 & 60.3 & 92.8 & 71.4 & 0.75 & 0.95 & 0.82 \\ \cline{2-2} \cline{4-11} %
	& \textcolor{color:naive-YOLOv3}{\rect} YOLO\,v3, na\"ive depth (COCO) \cite{Linder2018} & & 28 & 1081 & 68.7 & 93.1 & 74.7 & 0.80 & 0.92 & 0.84 \\ \cline{2-2} \cline{4-11} %
	& \textcolor{color:rgbd-YOLOv3}{\rect} RGB-D YOLO (COCO + synthetic 3D) \cite{Linder2020ICRA} & & 24 & 3269 & \underline{91.1} & \underline{99.0} & \underline{93.5} & \underline{0.92} & \underline{0.99} & \underline{0.94} \\ 
	 \hline
	2D, 3D, RGB-D & \begin{tabular}[c]{@{}l@{}} \textcolor{color:fusion3}{\ding{58}} Detection fusion\\[-1mm] DR-SPAAM + SECOND-DV (ILIAD) + RGB-D YOLO\end{tabular} & 360 & 9 & 9123 & - & - & - & 0.87 & 0.92 & 0.89 \\ \hline %
	3D, RGB-D & \begin{tabular}[c]{@{}l@{}} \textcolor{color:fusion2}{\ding{58}} Detection fusion\\[-1mm] SECOND-DV (ILIAD) + RGB-D YOLO \end{tabular} & 360 & 9 & 8294 & - & - & - & 0.91 & 0.96 & 0.93 \\ \hline %
	3D, RGB-D & \begin{tabular}[c]{@{}l@{}} \textcolor{color:fusion9}{\ding{58}} Detection fusion\\[-1mm] Same, but fuse 3D lidar only outside RGB-D FOV\end{tabular} & 360 & 9 & 8294 & - & - & - & \underline{0.92} & \underline{0.99} & \underline{0.94} \\ \hline %
\end{tabular}%

		}
		\caption{Cross-modal comparison of different human detectors on two distinct sequences from our data set in an industrial environment. Evaluation using 3D centroid annotations with a distance threshold of 0.5m is restricted to the Kinect v2 FOV for a fair comparison and performed on a 3.5 GHz Intel Quad-Core CPU with GTX 1080 GPU. ``ILIAD'' denotes our own intralogistics training data from the same environment.
      \label{fig:wp3-det-comparison-quantitative}
    }
    
\end{figure*}

Quantitative results of our cross-modal detector comparison are shown in \autoref{fig:wp3-det-comparison-quantitative}. For each of the two scenes, we present the precision-recall curves of all detectors in addition to the previously mentioned detection metrics.

\subsection{2D laser detectors}

Not surprisingly due to their sparseness, the approaches based on 2D laser data are overall the weakest methods in our evaluation. Our sensor mounting height close to the ground, which is required for 2D safety lasers, frequently leads to heavy occlusions by objects (e.g. pushcarts, brooms) or forks of other vehicles, partially explaining the weak results in the Dynamic scene (which additionally has people in challenging sitting and kneeing poses).
However, particularly on the less crowded Storage sequence, the strongest 2D laser methods can outperform all but the latest-generation 3D lidar and \mbox{RGB-D} detectors -- even without retraining on the target domain. This is remarkable and likely due to that scene being more favorable for detection in 2D laser, with subjects mostly in upright poses, and no other forklifts occluding the sensor.

We also notice that state-of-the-art deep learning approaches \cite{Beyer2018,Jia2020DRSPAAM} clearly outperform the earlier, classical method \cite{Arras2007} that only considers a single 2D laser scan, which is in line with results in the elderly care scenario of Beyer et al. \cite{Beyer2018}. However, the method by Leigh et al. \cite{Leigh2015}, which is a non-DL approach but like DR-SPAAM \cite{Jia2020DRSPAAM} considers temporal information\footnote{In the implementation by \cite{Jia2020DRSPAAM} that we used, DROW3x is run in a single-frame configuration (T=1) with non-maximum suppression instead of a voting grid to achieve fast real-time frame rates. It outperformed the original implementation \cite{Beyer2018} in terms of AP on our dataset. %
}, without retraining comes surprisingly close to both DL methods ($-2\%$ and $-4\%$ in peak-F1 over the combined sequences); for resource-constrained robots equipped with a 2D safety laser, this could be a reasonable choice over e.g. a resource-hungry DL-based detector if only medium-level detection accuracy is required. While the DL-based methods reach almost 50 Hz with GPU acceleration, our 2D safety laser provides raw data at only 12.5 Hz, thereby limiting the effective maximum detection rate.

\subsection{3D lidar-based approaches}

The 3D lidar detectors SECOND %
and its variant with dynamic voxelization, %
PointPillars,
Part A2-Net 
and the Object3D Detector %
suffer less from occlusion by other vehicles due to their higher vertical resolution compared to 2D laser, while being able to cover 360 degrees around the robot with a single sensor and detector instance\footnote{This clear benefit of the 3D lidar methods is not reflected in our metrics since we limit the evaluation to the smaller Kinect v2 FOV.}. However, the deeply-learned 3D approaches reach real-time performance at lower frame rates (7--11 Hz) than in 2D laser while at the same time requiring a more powerful GPU, whereas the Object3D Detector %
only requires a CPU to achieve a similar frame rate. %
Among all DL methods, PointPillars has the smallest GPU memory footprint, which makes it interesting for robotics applications which often deploy multiple networks in parallel. %

The performance of 3D lidar detectors is mostly consistent across both sequences except for the Object3D and PartA2-Net detectors. Both of these methods perform well in the open-space Dynamic scene in \autoref{fig:wp3-det-comparison-quantitative} (a), but their performance significantly degrades on the narrow Storage Room scene in \autoref{fig:wp3-det-comparison-quantitative} (b). The Object3D detector attains very low recall in the latter sequence due to the proposal generation through Euclidean clustering breaking down when persons are either close to a wall or to the handle bar of the ego-vehicle. In contrast, PartA2-Net retains the high recall, but loses precision due to an increased number of false positives. This could indicate that the method fails to generalize from open outdoor scenes to cluttered indoor environments.

SECOND achieves better detection results than PointPillars on both sequences, thereby demonstrating better generalization capabilities after training on KITTI. This is an interesting finding since PointPillars outperformed SECOND on both the KITTI and nuScenes \cite{Caesar2020CVPR} benchmarks. The results of SECOND are further improved by the addition of dynamic voxelization (SECOND-DV), which helps to preserve more information from the relatively sparse VLP-16 scans. This combination  leads to the second best performance among all of the modalities without additional training on the target domain. It falls short (by $-7.5\%$ in AP and $-6\%$ in peak-F1 score on the combined sequences) only to the results of the RGB-D YOLO method, which we use as a teacher during multi-sensor transfer learning on our target domain to further improve detection performance of SECOND-DV. These results will be discussed in \autoref{sec:transfer-learning-results-discussion}.

\subsection{RGB-D methods}
The RGB-D approaches %
are most limited in their of field of view, due to the Kinect v2 camera only covering around 86$^\circ$ horizontally. On the other hand, they can in theory provide detections at high sensor frame rates of up to 30 Hz. The latter does not apply in practice for RGB-D Pose 3D \cite{Zimmermann2018}, which runs only at single-digit FPS on a GPU in the publicly available implementation.

In comparison to other modalities and methods, \mbox{RGB-D Pose 3D}, YOLO\,v3 with na\"ive depth and \mbox{RGB-D YOLO} attain very high precision on the Storage Room scene. However, only \mbox{RGB-D YOLO} manages to achieve a similar level of performance also in the Dynamic sequence, which contains significant occlusion and persons frequently leaving and entering the field of view of the sensor. The na\"ive YOLO\,v3 detector provides inaccurate 3D localization under partial occlusion due to its inability to distinguish between foreground and background depth values within the detected 2D bounding box. \mbox{RGB-D Pose 3D}, PCL+HOG-SVM and Mobility Aids rely on point cloud data and therefore struggle with missing depth measurements at the image boundaries that lead to reduced recall. \mbox{RGB-D YOLO} implicitly combines both modalities in image space and is less affected by corruptions in either of them. %
It shows the best performance across all modalities and test sequences, which demonstrates the importance of both semantically rich RGB and geometrically accurate depth data. In terms of generalization, it is noteworthy that it has learned all 3D aspects (centroids, oriented bounding boxes) solely from synthetic training data \cite{Linder2019}, without any fine-tuning on real-world data from our target domain.

Qualitatively, the \mbox{RGB-D} methods appear to be the only ones that are robustly able to detect humans in any kind of body pose \emph{(e.g. sitting, kneeing, lying on the floor)}. Even though their limited field of view could be extended by combining several sensors, multiple instances of the Kinect~v2 driver alone would incur a very high GPU and CPU load, leading to very high power consumption that might be unsuitable for mobile use-cases. Due to the additional availability of visual appearance cues, having at least one \mbox{RGB(-D)} sensor onboard could be valuable for service robotics use-cases that require person re-identification capabilities, for instance for person guidance or following \cite{Linder2016}. At the same time, privacy concerns should be taken into consideration, which are less of a problem with 2D laser and 3D lidar.

\subsection{Multi-sensor transfer learning results}
\label{sec:transfer-learning-results-discussion}

We now compare the domain-adapted SECOND-DV 3D lidar detector, which we trained completely from scratch on data from the target domain using our multi-sensor transfer learning approach, to the earlier discussed variant trained on the KITTI dataset. We can see in Fig.~ \ref{fig:wp3-det-comparison-quantitative} that even with our rather simplistic approach without tracking, due to using a strong \mbox{RGB-D} detector as teacher, we obtain very promising results for SECOND-DV that outperform the original model trained on hand-labeled KITTI data in terms of AP on both scenes (by up to $+11.5\%$, or $+6.9\%$ combined) and in peak-F1 score on the second scene ($+7\%$), without having manually annotated a single lidar point cloud. Weak supervision led to a significant improvement in AP due to higher maximum recall, with a slight drop in precision only in the Dynamic scene. There, the adapted lidar model achieves an even higher recall than its teacher network.
The increased peak-F1 score on the Storage Room sequence suggests a better adaptation to cluttered indoor scenes.

\subsection{Detection fusion of best-performing detectors}

As can be seen in the last three rows of \autoref{fig:wp3-det-comparison-quantitative}, na\"ively combining detections from multiple sensors can improve recall, but at the cost of reduced precision as the weaker sensor/detector combinations introduce additional false positives. Therefore, more complex voting schemes or detection-to-track fusion (after calibrating detector scores) should be utilized. Under the current setup, when using a strong detector in combination with a time-of-flight \mbox{RGB-D} camera, it appears reasonable to fuse 3D lidar only outside the \mbox{RGB-D} camera's field of view for 360-degree coverage.

\section{Conclusion}

In this paper, we performed a cross-modal analysis of human detection approaches with experimental focus on industrial environments -- an important robotics use-case that is underrepresented in current benchmarks. %
We also showed that already a relatively simple multi-sensor transfer learning approach can effectively address training data scarcity in our target domain, and improve detector performance.

The main conclusions from our experiments are as follows: Recent state-of-the-art detectors, which have been developed and tested on other domains (e.g. automotive, elderly care facilities), are in general also the top-performing ones in our industrial target domain.
In particular, strong \mbox{RGB-D} methods 
work well even when learning 3D localization just from synthetic data, and have no problems with detecting persons in unusual protective clothing if pre-trained on large-scale, real-world 2D image datasets such as MS COCO.
Instead, 3D lidar-based approaches show a larger domain gap that can be mitigated by retraining on data from the target domain -- for example using the presented transfer learning approach. There, it is noteworthy that if we learn 3D localization and bounding box estimation in RGB-D YOLO from \emph{synthetic} data, after multi-sensor transfer learning we obtain a 3D lidar detector model that outperforms other models (from a different domain) on \emph{real-world} data.
The domain gap that exists without domain adaptation suggests that currently available public datasets for 3D lidar are too small or not diverse enough to make deep learning approaches generalize well to robotics scenarios like ours.

Finally, we want to revisit the two example robots from Section I from a practical point of view: For the cost-sensitive consumer robot without GPU in a moderately complex use-case, a very promising combination appears to be a (low-cost) 2D lidar and the classical ML approach by Leigh \emph{et al.} \cite{Leigh2015}. In our experiments, its performance was close to the best deep learning approach for this modality and surprisingly decent in comparison to the other much richer sensors.
For the hospital delivery robot, a strong drop in performance from RGB-D to 3D lidar was not confirmed in our experiments. The combination of an RGB-D camera with our approach \cite{Linder2020ICRA} was the best one overall, but relatively closely followed by 3D lidar and SECOND-DV, if retrained on the target domain using the presented transfer learning strategy. If privacy were a concern in the use-case, therefore 3D lidar might be the preferred choice.

These findings point into several directions for future work. %
The observed generalization issues of the 3D lidar detectors, which apparently suffer more from a deployment gap than RGB-D methods, highlight the importance of dedicated data sets in robotics applications domains. Along that  line, we will consider the JackRabbot data set \cite{MartinMartin2019} and a re-evaluation as soon as data with next-generation sensors becomes available, such as the Azure Kinect.
A more profound exploration of detection fusion also appears a promising direction.

\footnotesize{
\section*{ACKNOWLEDGMENT}

We would like to thank Manuel Fernandez Carmona, Kevin Li Sun, Martin Magnusson and the entire ILIAD team for help with data recording.
}

\bibliographystyle{IEEEtran}
\bibliography{main}

\begin{thebibliography}{10}
\providecommand{\url}[1]{#1}
\csname url@rmstyle\endcsname
\providecommand{\newblock}{\relax}
\providecommand{\bibinfo}[2]{#2}
\providecommand\BIBentrySTDinterwordspacing{\spaceskip=0pt\relax}
\providecommand\BIBentryALTinterwordstretchfactor{4}
\providecommand\BIBentryALTinterwordspacing{\spaceskip=\fontdimen2\font plus
\BIBentryALTinterwordstretchfactor\fontdimen3\font minus
  \fontdimen4\font\relax}
\providecommand\BIBforeignlanguage[2]{{%
\expandafter\ifx\csname l@#1\endcsname\relax
\typeout{** WARNING: IEEEtran.bst: No hyphenation pattern has been}%
\typeout{** loaded for the language `#1'. Using the pattern for}%
\typeout{** the default language instead.}%
\else
\language=\csname l@#1\endcsname
\fi
#2}}

\bibitem{Linder2020ICRA}
T.~Linder, K.~Y. Pfeiffer, N.~Vaskevicius, R.~Schirmer, and K.~O. Arras,
  ``Accurate detection and {3D} localization of humans using a novel
  {YOLO}-based {RGB-D} fusion approach and synthetic training data,'' in
  \emph{{IEEE} Int.~Conf.~on Robotics and Automation {(ICRA)}}, 2020.

\bibitem{Bellotto2018}
N.~Bellotto, S.~Cosar, and Z.~Yan, ``Human detection and tracking,''
  \emph{Encyclopedia of Robotics}, 2018.

\bibitem{Ciaparrone2019}
G.~Ciaparrone, F.~L. Sánchez, S.~Tabik, L.~Troiano, R.~Tagliaferri, and
  F.~Herrera, ``Deep learning in video multi-object tracking: A survey,''
  \emph{Neurocomputing}, 2019.

\bibitem{Bellotto2009}
N.~Bellotto and H.~Hu, ``Multisensor-based human detection and tracking for
  mobile service robots,'' \emph{IEEE Transactions on Systems, Man, and
  Cybernetics, Part B (Cybernetics)}, 2009.

\bibitem{Linder2016}
T.~Linder, S.~Breuers, B.~Leibe, and K.~O. Arras, ``On multi-modal people
  tracking from mobile platforms in very crowded and dynamic environments,'' in
  \emph{{IEEE} Int.~Conf.~on Robotics and Automation {(ICRA)}}, 2016.

\bibitem{Volkhardt2013}
M.~Volkhardt, C.~Weinrich, and H.~M. Gross, ``Multi-modal people tracking on a
  mobile companion robot,'' in \emph{European Conference on Mobile Robotics
  {(ECMR)}}, 2013.

\bibitem{Qi2018}
C.~R. Qi, W.~Liu, C.~Wu, H.~Su, and L.~J. Guibas, ``Frustum {P}oint{N}ets for
  {3D} object detection from {RGB-D} data,'' in \emph{{IEEE} Conf. on Computer
  Vision and Pattern Recognition {(CVPR)}}, 2018.

\bibitem{ku2018}
J.~Ku, M.~Mozifian, J.~Lee, A.~Harakeh, and S.~Waslander, ``Joint {3D} proposal
  generation and object detection from view aggregation,'' in \emph{{IEEE/RSJ}
  Int.~Conf.~on Intelligent Robots and Systems {(IROS)}}, 2018.

\bibitem{luber2011}
M.~Luber, G.~D. Tipaldi, and K.~O. Arras, ``Place-dependent people tracking,''
  \emph{Int.~Journal of Robotics Research {(IJRR)}}, 2011.

\bibitem{Weinrich2014}
C.~Weinrich, T.~Wengefeld, C.~Schr{\"{o}}ter, and H.~Gross, ``People detection
  and distinction of their walking aids in {2D} laser range data based on
  generic distance-invariant features,'' in \emph{IEEE Int~.Symposium on Robot
  and Human Interactive Communication {(RO-MAN)}}, 2014.

\bibitem{Beyer2018}
L.~Beyer, A.~Hermans, T.~Linder, K.~O. Arras, and B.~Leibe, ``Deep person
  detection in two-dimensional range data,'' \emph{{IEEE} Robotics and
  Automation Letters}, 2018.

\bibitem{Geiger2012}
A.~Geiger, P.~Lenz, and R.~Urtasun, ``Are we ready for autonomous driving?
  {T}he {KITTI} vision benchmark suite,'' in \emph{{IEEE} Conf. on Computer
  Vision and Pattern Recognition {(CVPR)}}, 2012.

\bibitem{Sun2020CVPR}
P.~Sun, H.~Kretzschmar, X.~Dotiwalla, A.~Chouard, V.~Patnaik, P.~Tsui, J.~Guo,
  Y.~Zhou, Y.~Chai, B.~Caine, V.~Vasudevan, W.~Han, J.~Ngiam, H.~Zhao,
  A.~Timofeev, S.~Ettinger, M.~Krivokon, A.~Gao, A.~Joshi, Y.~Zhang, J.~Shlens,
  Z.~Chen, and D.~Anguelov, ``Scalability in perception for autonomous driving:
  Waymo open dataset,'' in \emph{{IEEE} Conf. on Computer Vision and Pattern
  Recognition {(CVPR)}}, 2020.

\bibitem{Caesar2020CVPR}
H.~Caesar, V.~Bankiti, A.~H. Lang, S.~Vora, V.~E. Liong, Q.~Xu, A.~Krishnan,
  Y.~Pan, G.~Baldan, and O.~Beijbom, ``nu{S}cenes: A multimodal dataset for
  autonomous driving,'' in \emph{{IEEE} Conf. on Computer Vision and Pattern
  Recognition {(CVPR)}}, 2020.

\bibitem{MartinMartin2019}
R.~Martín-Martín, H.~Rezatofighi, A.~Shenoi, M.~Patel, J.~Gwak, N.~Dass,
  A.~Federman, P.~Goebel, and S.~Savarese, ``{JRDB}: A dataset and benchmark
  for visual perception for navigation in human environments,'' \emph{arXiv
  preprint}, 2019.

\bibitem{jia2021domaingaps}
D.~Jia, A.~Hermans, and B.~Leibe, ``Domain and modality gaps for {LiDAR}-based
  person detection on mobile robots,'' \emph{ar{X}iv preprint}, 2021.

\bibitem{rist2019iv}
C.~B. {Rist}, M.~{Enzweiler}, and D.~M. {Gavrila}, ``Cross-sensor deep domain
  adaptation for {LiDAR} detection and segmentation,'' in \emph{IEEE
  Intelligent Vehicles Symposium (IV)}, 2019.

\bibitem{piewak2019itsc}
F.~{Piewak}, P.~{Pinggera}, and M.~{Zöllner}, ``Analyzing the cross-sensor
  portability of neural network architectures for {LiDAR}-based semantic
  labeling,'' in \emph{{IEEE} Intelligent Transportation Systems Conference
  (ITSC)}, 2019.

\bibitem{piewak2018eccv}
F.~Piewak, P.~Pinggera, M.~Sch{\"a}fer, D.~Peter, B.~Schwarz, N.~Schneider,
  M.~Enzweiler, D.~Pfeiffer, and M.~Z{\"o}llner, ``Boosting {LiDAR}-based
  semantic labeling by cross-modal training data generation,'' in
  \emph{Computer Vision -- ECCV 2018 Workshops}, 2018.

\bibitem{wang2019ral}
B.~H. {Wang}, W.~{Chao}, Y.~{Wang}, B.~{Hariharan}, K.~Q. {Weinberger}, and
  M.~{Campbell}, ``{LDLS}: 3-{D} object segmentation through label diffusion
  from 2-{D} images,'' \emph{{IEEE} Robotics and Automation Letters}, 2019.

\bibitem{Yan2018}
Z.~Yan, L.~Sun, T.~Duckett, and N.~Bellotto, ``Multisensor online transfer
  learning for 3d lidar-based human detection with a mobile robot,'' in
  \emph{{IEEE/RSJ} Int.~Conf.~on Intelligent Robots and Systems {(IROS)}},
  2018.

\bibitem{Arras2007}
K.~O. Arras, O.~M. Mozos, and W.~Burgard, ``Using boosted features for the
  detection of people in {2D} range data,'' in \emph{{IEEE} Int.~Conf.~on
  Robotics and Automation {(ICRA)}}, 2007.

\bibitem{Leigh2015}
A.~Leigh, J.~Pineau, N.~Olmedo, and H.~Zhang, ``Person tracking and following
  with 2d laser scanners,'' in \emph{{IEEE} Int.~Conf.~on Robotics and
  Automation {(ICRA)}}, 2015.

\bibitem{Jia2020DRSPAAM}
D.~Jia, A.~Hermans, and B.~Leibe, ``{DR-SPAAM}: A spatial-attention and
  auto-regressive model for person detection in 2{D} range data,'' in
  \emph{{IEEE/RSJ} Int.~Conf.~on Intelligent Robots and Systems {(IROS)}},
  2020.

\bibitem{MMDetection3D}
\BIBentryALTinterwordspacing
MMDetection3D, ``Open-source 3{D} object detection toolbox by {OpenMMLab},''
  Multimedia Lab at the Chinese University of Hong Kong, 2020. [Online].
  Available: \url{https://github.com/open-mmlab/mmdetection3d}
\BIBentrySTDinterwordspacing

\bibitem{yan2018sensors}
Y.~Yan, Y.~Mao, and B.~Li, ``{SECOND}: Sparsely {E}mbedded {C}onvolutional
  {D}etection,'' \emph{Sensors}, 2018.

\bibitem{Zhou2019endtoend}
Y.~Zhou, P.~Sun, Y.~Zhang, D.~Anguelov, J.~Gao, T.~Ouyang, J.~Guo, J.~Ngiam,
  and V.~Vasudevan, ``End-to-end multi-view fusion for 3d object detection in
  {LiDAR} point clouds,'' in \emph{Conference on Robot Learning {(CoRL)}},
  2020.

\bibitem{Lang2019}
A.~H. Lang, S.~Vora, H.~Caesar, L.~Zhou, J.~Yang, and O.~Beijbom,
  ``Point{P}illars: {F}ast {E}ncoders for {O}bject {D}etection from {P}oint
  {C}louds,'' in \emph{{IEEE} Conf. on Computer Vision and Pattern Recognition
  {(CVPR)}}, 2019.

\bibitem{shi2020part}
S.~Shi, Z.~Wang, J.~Shi, X.~Wang, and H.~Li, ``From {P}oints to {P}arts: 3{D}
  {O}bject {D}etection from {P}oint {C}loud with {P}art-aware and
  {P}art-aggregation {N}etwork,'' \emph{{IEEE} Transactions on Pattern Analysis
  and Machine Intelligence {(PAMI)}}, 2020.

\bibitem{Zhou2018}
Y.~Zhou and O.~Tuzel, ``Voxel{N}et: End-to-end learning for point cloud based
  3{D} object detection,'' in \emph{{IEEE} Conf. on Computer Vision and Pattern
  Recognition {(CVPR)}}, 2018.

\bibitem{Liu2016}
W.~Liu, D.~Anguelov, D.~Erhan, C.~Szegedy, S.~Reed, C.-Y. Fu, and A.~C. Berg,
  ``{SSD}: Single shot multibox detector,'' in \emph{European Conference on
  Computer Vision {(ECCV)}}, 2016.

\bibitem{COCO14}
T.-Y. Lin, M.~Maire, S.~Belongie, J.~Hays, P.~Perona, D.~Ramanan,
  P.~Doll{\'a}r, and C.~L. Zitnick, ``Microsoft {C}{O}{C}{O}: {C}ommon
  {O}bjects in {C}ontext,'' in \emph{European Conference on Computer Vision
  {(ECCV)}}, 2014.

\bibitem{Linder2019}
T.~Linder, M.~J. Hernandez~Leon, N.~Vaskevicius, and K.~O. Arras, ``Towards
  training person detectors for mobile robots using synthetically generated
  {RGB-D} data,'' in \emph{Computer Vision and Pattern Recognition ({CVPR})
  2019 Workshop on 3{D} Scene Generation}, 2019.

\bibitem{Linder2018}
T.~Linder, D.~Griesser, N.~Vaskevicius, and K.~O. Arras, ``Towards accurate
  {3D} person detection and localization from {RGB-D} in cluttered
  environments,'' in \emph{IROS 2018 Workshop on Robotics for Logistics in
  Warehouses and Environments Shared with Humans}, 2018.

\bibitem{Munaro2014}
M.~Munaro and E.~Menegatti, ``Fast {RGB-D} people tracking for service
  robots,'' \emph{Autonomous Robots (AURO)}, 2014.

\bibitem{Vasquez2017}
A.~Vasquez, M.~Kollmitz, A.~Eitel, and W.~Burgard, ``Deep detection of people
  and their mobility aids for a hospital robot,'' in \emph{European Conference
  on Mobile Robotics {(ECMR)}}, 2017.

\bibitem{Zimmermann2018}
C.~Zimmermann, T.~Welschehold, C.~Dornhege, W.~Burgard, and T.~Brox, ``{3D}
  human pose estimation in {RGBD} images for robotic task learning,'' in
  \emph{Proc. {IEEE} Int. Conf. on Robotics and Automation {(ICRA)}}, 2018.

\bibitem{Cao2017}
Z.~Cao, T.~Simon, S.-E. Wei, and Y.~Sheikh, ``Realtime multi-person 2{D} pose
  estimation using part affinity fields,'' in \emph{{IEEE} Conf. on Computer
  Vision and Pattern Recognition {(CVPR)}}, 2017.

\bibitem{jia2020selfsupervised}
D.~Jia, M.~Steinweg, A.~Hermans, and B.~Leibe, ``{Self-Supervised Person
  Detection in 2D Range Data using a Calibrated Camera},'' in \emph{{IEEE}
  Int.~Conf.~on Robotics and Automation {(ICRA)}}, 2021.

\bibitem{Beyer2015}
L.~Beyer, A.~Hermans, and B.~Leibe, ``Biternion nets: Continuous head pose
  regression from discrete training labels,'' \emph{Pattern Recognition: 37th
  German Conference, (GCPR)}, 2015.

\bibitem{Lewandowski2019a}
B.~Lewandowski, D.~Seichter, T.~Wengefeld, L.~Pfennig, H.~Drumm, and H.-M.
  Gross, ``{Deep Orientation: Fast and Robust Upper Body Orientation Estimation
  for Mobile Robotic Applications},'' in \emph{{IEEE/RSJ} Int.~Conf.~on
  Intelligent Robots and Systems {(IROS)}}, 2019.

\end{thebibliography}

\end{document}